\documentclass{article}

\usepackage[preprint]{nips_2018}
\usepackage{amsmath}
\usepackage{natbib}
\usepackage{tabu,multirow}
 \usepackage{changepage}
\usepackage{makecell}

\usepackage[utf8]{inputenc} 
\usepackage[T1]{fontenc}    
\usepackage{hyperref}       
\usepackage{url}            
\usepackage{booktabs}       
\usepackage{amsfonts}       
\usepackage{nicefrac}       
\usepackage{microtype}      
\newcommand{\ignore}[1]{}
\usepackage{graphicx}
\usepackage{subfigure}
\title{Attentioned Convolutional LSTM Inpainting Network for Anomaly Detection in Videos }

\author{
  Itamar Ben-Ari \\
  Advanced Analytics\\
  Intel\\
  \texttt{itamar.ben-ari@intel.com} \\
  \And
  Ravid Shwartz-Ziv \thanks{both authors contributed equally} \\
  Advanced Analytics\\
  Intel\\
  \texttt{ravid.ziv@intel.com} \\
}

\begin{document}

\maketitle

\begin{abstract}

We propose a semi-supervised model for detecting anomalies in videos inspired by the Video Pixel Network \citep{van2016conditional}. VPN is a probabilistic generative model based on a deep neural network that estimates the discrete joint distribution of raw pixels in video frames. Our model extends the Convolutional-LSTM video encoder part of the VPN with a novel convolutional based attention mechanism. We also modify the Pixel-CNN decoder part of the VPN to a frame inpainting task where a partially masked version of the frame to predict is given as input. The frame reconstruction error is used as an anomaly indicator. We test our model on a modified version of the moving mnist dataset \citep{srivastava2015unsupervised}. Our model is shown to be effective in detecting anomalies in videos. This approach could be a component in applications requiring visual common sense.

\end{abstract}
\section{\label{Introduction}Introduction}

\begin{figure}
\subfigure[Network topology] {%
\label{fig:model}%
\includegraphics[height=2.5in]{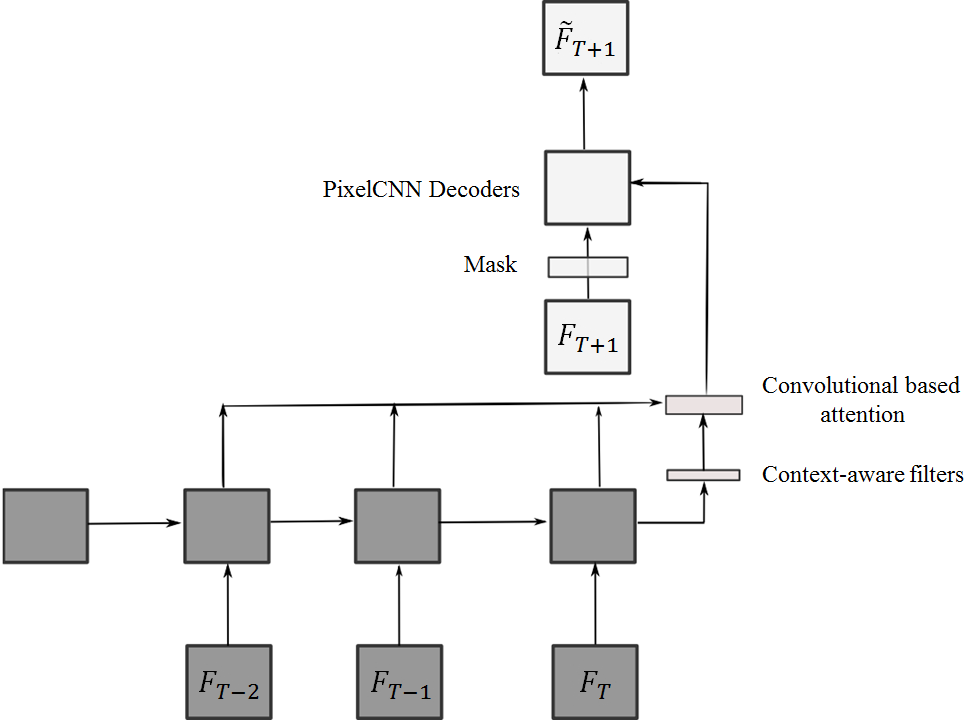}}%
\subfigure[Masked frame] {%
\label{fig:mask_example}%
\includegraphics[height=2.5in]{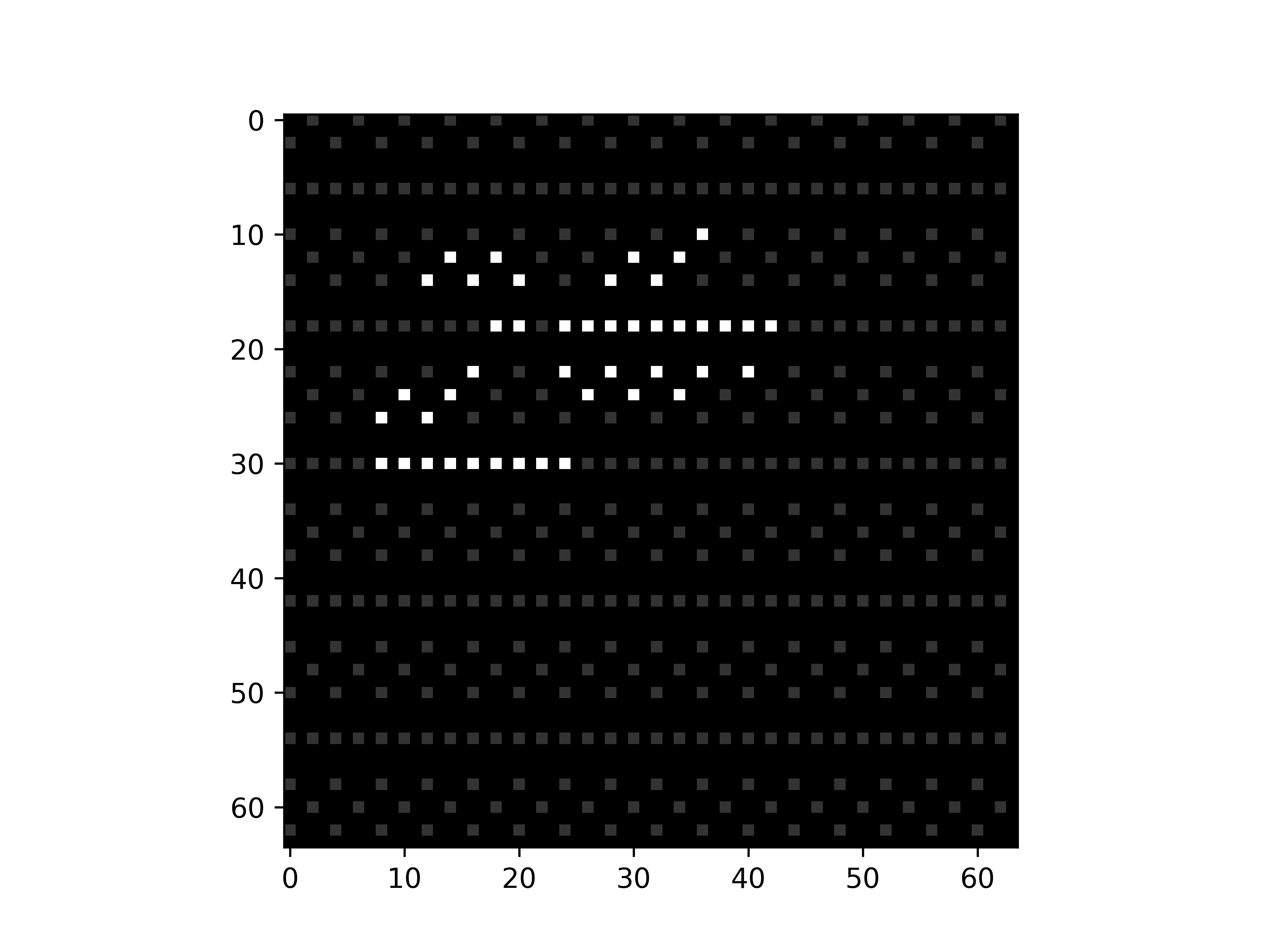}}%
\end{figure}

Real-time anomaly detection in videos has significant value across many domains such as robot patrolling \citep{chakravarty2007anomaly} and visual inspection of manufacturing processes. The task remains challenging due to the complexity and variability of the data and high computational cost in an edge device setting. Current approaches range from supervised models based on Convolutional Neural Networks (CNNs) architectures \citep{sabokrou2018deep}, long-term temporal dynamic models such as Recurrent Neural Networks (RNNs) \citep{radford2018network} and unsupervised models for video features learning \citep{zhang2016combining, pham2011detection, zhao2011online}. In this paper we propose an encoder-decoder network where the input is a sequence of frames with the last frame partially masked and the output is a reconstruction of that frame. We use the reconstruction error as an indicator for an anomaly in the sequence, where we assume the model will reconstruct a masked pixel with a typical value, in accordance with that pixel’s spatial and temporal context. A pixel containing unexpected / out of context value will be poorly reconstructed and indicate an anomaly. Our model is inspired by the Video Pixel Networks (VPN) \citep{kalchbrenner2016video} with two main differences - 1. We add a convolutional based attention mechanism where the filters weights are dynamic and input dependent. This mechanism utilizes the local structure of images better than a standard global weighting attention mechanism. 2. A partially masked version of the frame to predict is given as input to the model (see figure \ref{fig:mask_example}) - this eliminates the need for masked convolutions and enables the computation of the predicted distribution of all pixels to be parallelized. The model can also utilize information from an unmasked neighborhood of a predicted pixel which makes the prediction task tractable. Using the proposed modifications above our model is able to find anomalies in videos in an unsupervised manner and in real-time.

\section{Related work}
\subsection{Anomaly Detection}

Unsupervised and semi-supervised video anomaly detection models can be classified into three main categories - 
1. Representation learning for reconstruction: Encoder-Decoder Methods which transform the input into a hidden representation and then try to reconstruct it. Anomalies are represented by poorly reconstructed deviations from the source. Principal Component Analysis (PCA) and Auto-encoders (AEs) are examples of such models.
2. Predictive modeling: where the sequence of frames is viewed as a time series and the model’s task is to predict the next frame pixels values distribution. Anomalies are represented by pixels with low likelihood values. Auto-Regressive models and Convolutional-LSTMs are examples of such models.
3) Generative models: e.g., Generative Adversarial Networks (GAN) and Variational Auto-Encoders (VAE), which can compute a measure of frame abnormality.

\subsection{Video Pixel Network}
VPN is a frame predictive model shown to give SOTA results on the moving mnist and pushing robots datasets \citep{finn2016unsupervised}. The architecture of the VPN consists of two parts: A CNN resolution preserving encoder and a Pixel-CNN decoder \citep{van2016conditional}. The CNN encoder output is aggregated over time by a Convolutional-LSTM in order to capture temporal dependencies. The Pixel-CNN decoder uses masked convolutions to model space and color dependencies in the predicted frame (by allowing a flow of information from previously predicted pixels to a current predicted pixel). The last layer of the Pixel-CNN decoder is a softmax layer over 256 intensity values for each color channel in each pixel.  
\section{Our model}

\subsection{Convolutional based attention mechanism}
The relevant context window for frame prediction may vary in size and frames importance distribution. An attention mechanism is a popular tool used to overcome memory limitations of recurrent models and bring to focus relevant parts of a context window. Since current attention mechanisms do not leverage the local structure of images, we propose the use of a convolution with input dependent filter weights to generate an attention like mechanism \citep{atten2018}. We use a small meta-network to output context-sensitive convolution filters, which are then applied to a tensor of concatenated Convolutional-LSTM outputs (representing the context window). The Convolutional-LSTM and convolutional attention output tensors preserve the spatial dimensions and local structure of the video frames. This allows us to concatenate the partially masked frame as an additional channel of the attention output tensor and forward it to the inpainting network for reconstruction (see figure \ref{fig:model}).     

\subsection{Convolutions with masked frames for image inpainting}
In the VPN model the frame to be predicted is given as input in the training phase. The PixelCNN decoder uses masked convolutions to ensure the predicted pixel does not "see" its label (i.e. true value). The masked convolution only uses information from pixels preceding the predicted pixel (for a top-bottom left-right pixel order), enabling the network to model some of the spatial dependencies in the predicted frame. In inference time the pixels are predicted sequentially. In our anomaly detection reconstruction approach the frame to be predicted is also given as input but is partially masked, blocking the flow of information from a label to a masked pixel. The modeling of spatial dependencies of a pixel is enabled by using information from non-masked pixels in its neighborhood. We use a grid mask with random shifts where the portion of masked pixels in the frame is $\sim$ 95\% (see figure \ref{fig:mask_example}). This way the model learns a general structure of the frame and must rely on temporal dependencies. In inference time the same procedure is applied, so the pixels are predicted in parallel, resulting in real-time detection. 

\subsection{Loss function as an anomaly measure}
We use the log-likelihood of the pixels values given the network predicted distribution as a loss function.  The average pixels log-likelihood is used as a global score for frame abnormality, where we assume the pixels are independently distributed given the unmasked pixels and context window frames. The loss is defined as:
\begin{equation} \label{eq1}
\begin{split}
L(X_T) &= -\log{P\left(X_{T}|\Tilde{X}_{T}, X_{<T}, \theta \right)}
    =-\log{\prod{P_{i,j,c}\left(X_{T}|\Tilde{X}_{T},X_{<T}, \theta\right)}} \\ &
    =-\sum_{i,j}\log{(y_{i,j,R}[X_{i,j,R}])} + \log{(y_{i,j,G}[X_{i,j,G}])} + \log{(y_{i,j,B}[X_{i,j,B}])}
\end{split}
\end{equation}
where ${X}_{T}$ are the pixels of the frame to reconstruct in time ${T}$, $\Tilde{X}_{T}$ is the masked frame, $X_{<T}$ are all the frames prior to the $T$-th frame, $\theta$ are the network parameters, $X_{i,j,c}$ is the value of channel $c$ of the $i,j$ pixel of frame ${X}_{T}$ and $y_{i,j,c}$ is the predicted distribution for that value.

In the training phase we train our network only on anomaly free videos. This way the network learns to predict a distribution for pixel values showing normal behavior, and will give low probability predictions for abnormal values in inference time. We use the log-likelihood as an anomaly measure where low likelihood pixel values indicate higher chance for these pixels to show an anomaly. 

\section{Experiments}
We evaluate each contribution proposed in this paper, convolutional-based attention and masked frame reconstruction, on a modified version of the Moving MNIST dataset \citep{srivastava2015unsupervised}. We show that our model can learn both the temporal and spatial aspects of the movies and automatically detect anomalies without explicit supervision. We compare two methods as baselines: the original VPN model and Conv-LSTM network (which detects abnormal frames based on the reconstruction error \citep{medel2016anomaly}), together with two variants of our model - the first omits the masked frame from the input and the second does not use attention.

\begin{figure}
\label{fig:anomaly_example}%
\centering
\includegraphics[height=2.5in]{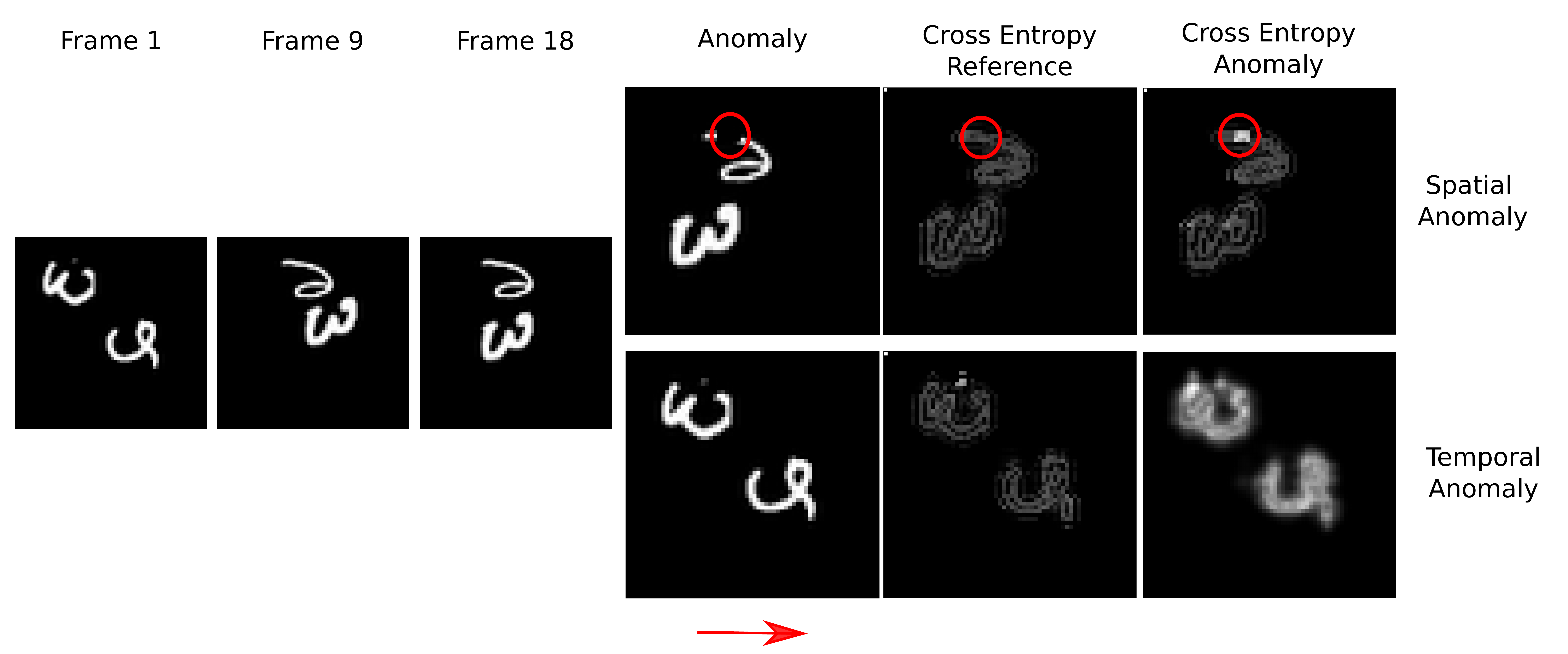}
\caption{Example of a corrupted moving MNIST sequence. On the left there are 3 un-corrupted frames from the beginning of the sequence. On the right, a black square corruption (top) and a temporal corruption (bottom), along with their prediction loss map (for both the corrupted frames and their un-corrupted version). The bright values in the loss maps correspond to high cross-entropy loss values indicating an anomaly.}
\end{figure}

\textbf{Dataset} - The Moving MNIST is a common dataset consisting of two digits moving independently in a frame (potentially overlapping) with constant velocity. It consists of sequences of 20 frames of size $64\times64$. The training sequences are generated on-the-fly by sampling MNIST digits and generating trajectories with randomly sampled velocity and angle. The training set was downloaded from \citep{srivastava2015unsupervised} and consists of 10000 sequences. Our test set consists of both normal and corrupted sequences. In order to generate a corrupted sequence, we replace the last frame with the first frame and paint a "corruption" of $3\times3$ black pixels on a digit (see figure \ref{fig:anomaly_example}). These corruptions are in two dimensions - temporal (changing the frame order) and spatial (painting the black square).

\textbf{Evaluation Metric} - We use the Equal Error Rate (EER) which is the accuracy value for equal precision and recall, a standard metric in abnormal event detection.

\textbf{Results} - Table \ref{results_table} shows the EER for the different models tested. Our model outperforms both the baseline models (VPN and Conv-LSTM) and the partial variations of our model, showing the importance of each contribution. Replacing the last frame with the first tests the ability of the models to detect temporal anomalies in the sequence. In such anomalies the attention mechanism improves the model’s ability to capture the abnormal frame-to-frame changes. The black square corruption tests the ability of the model to capture spatial dependencies. Our masked frame approach captures the dependencies between a masked pixel and its unmasked neighborhood, resulting in the reconstruction of the original values of the blackened pixels, i.e. predicting low probability for zero values.

\begin{table}[h!]
\label{results_table}
\begin{center}
\begin{tabular}{l|c}
\Xhline{2\arrayrulewidth}

\textbf{Model}                                      & \textbf{EER [\%]} \\
\Xhline{1\arrayrulewidth}

Conv-LSTM                                  & 70.3    \\
VPN                                        & 82.1   \\ 
Our model                                  & 87   \\
Our model w/o the masked convolutions      & 84.6  \\
Our model w/o the conv-attention mechanism & 85.7  \\
\Xhline{2\arrayrulewidth}
\end{tabular}
\end{center}
\end{table}

\bibliography{ref}

\begin{thebibliography}{12}
\providecommand{\natexlab}[1]{#1}
\providecommand{\url}[1]{\texttt{#1}}
\expandafter\ifx\csname urlstyle\endcsname\relax
  \providecommand{\doi}[1]{doi: #1}\else
  \providecommand{\doi}{doi: \begingroup \urlstyle{rm}\Url}\fi

\bibitem[van~den Oord et~al.(2016)van~den Oord, Kalchbrenner, Espeholt,
  Vinyals, Graves, et~al.]{van2016conditional}
Aaron van~den Oord, Nal Kalchbrenner, Lasse Espeholt, Oriol Vinyals, Alex
  Graves, et~al.
\newblock Conditional image generation with pixelcnn decoders.
\newblock In \emph{Advances in Neural Information Processing Systems}, pages
  4790--4798, 2016.

\bibitem[Srivastava et~al.(2015)Srivastava, Mansimov, and
  Salakhudinov]{srivastava2015unsupervised}
Nitish Srivastava, Elman Mansimov, and Ruslan Salakhudinov.
\newblock Unsupervised learning of video representations using lstms.
\newblock In \emph{International conference on machine learning}, pages
  843--852, 2015.

\bibitem[Chakravarty et~al.(2007)Chakravarty, Zhang, Jarvis, and
  Kleeman]{chakravarty2007anomaly}
Punarjay Chakravarty, Alan~M Zhang, Ray Jarvis, and Lindsay Kleeman.
\newblock Anomaly detection and tracking for a patrolling robot.
\newblock In \emph{Australasian Conference on Robotics and Automation (ACRA)}.
  Citeseer, 2007.

\bibitem[Sabokrou et~al.(2018)Sabokrou, Fayyaz, Fathy, Moayed, and
  Klette]{sabokrou2018deep}
Mohammad Sabokrou, Mohsen Fayyaz, Mahmood Fathy, Zahra Moayed, and Reinhard
  Klette.
\newblock Deep-anomaly: Fully convolutional neural network for fast anomaly
  detection in crowded scenes.
\newblock \emph{Computer Vision and Image Understanding}, 2018.

\bibitem[Radford et~al.(2018)Radford, Apolonio, Trias, and
  Simpson]{radford2018network}
Benjamin~J Radford, Leonardo~M Apolonio, Antonio~J Trias, and Jim~A Simpson.
\newblock Network traffic anomaly detection using recurrent neural networks.
\newblock \emph{arXiv preprint arXiv:1803.10769}, 2018.

\bibitem[Zhang et~al.(2016)Zhang, Lu, Zhang, and Ruan]{zhang2016combining}
Ying Zhang, Huchuan Lu, Lihe Zhang, and Xiang Ruan.
\newblock Combining motion and appearance cues for anomaly detection.
\newblock \emph{Pattern Recognition}, 51:\penalty0 443--452, 2016.

\bibitem[Pham et~al.(2011)Pham, Saha, Phung, and Venkatesh]{pham2011detection}
Duc~Son Pham, Budhaditya Saha, Dinh~Q Phung, and Svetha Venkatesh.
\newblock Detection of cross-channel anomalies from multiple data channels.
\newblock In \emph{Data Mining (ICDM), 2011 IEEE 11th International Conference
  on}, pages 527--536. IEEE, 2011.

\bibitem[Zhao et~al.(2011)Zhao, Fei-Fei, and Xing]{zhao2011online}
Bin Zhao, Li~Fei-Fei, and Eric~P Xing.
\newblock Online detection of unusual events in videos via dynamic sparse
  coding.
\newblock In \emph{Computer Vision and Pattern Recognition (CVPR), 2011 IEEE
  Conference on}, pages 3313--3320. IEEE, 2011.

\bibitem[Kalchbrenner et~al.(2016)Kalchbrenner, Oord, Simonyan, Danihelka,
  Vinyals, Graves, and Kavukcuoglu]{kalchbrenner2016video}
Nal Kalchbrenner, Aaron van~den Oord, Karen Simonyan, Ivo Danihelka, Oriol
  Vinyals, Alex Graves, and Koray Kavukcuoglu.
\newblock Video pixel networks.
\newblock \emph{arXiv preprint arXiv:1610.00527}, 2016.

\bibitem[Finn et~al.(2016)Finn, Goodfellow, and Levine]{finn2016unsupervised}
Chelsea Finn, Ian Goodfellow, and Sergey Levine.
\newblock Unsupervised learning for physical interaction through video
  prediction.
\newblock In \emph{Advances in neural information processing systems}, pages
  64--72, 2016.

\bibitem[Shen et~al.(2017)Shen, Min, Li, and Carin]{atten2018}
Dinghan Shen, Martin~Renqiang Min, Yitong Li, and Lawrence Carin.
\newblock Adaptive convolutional filter generation for natural language
  understanding.
\newblock \emph{CoRR}, abs/1709.08294, 2017.
\newblock URL \url{http://arxiv.org/abs/1709.08294}.

\bibitem[Medel and Savakis(2016)]{medel2016anomaly}
Jefferson~Ryan Medel and Andreas Savakis.
\newblock Anomaly detection in video using predictive convolutional long
  short-term memory networks.
\newblock \emph{arXiv preprint arXiv:1612.00390}, 2016.

\end{thebibliography}

\end{document}